\documentclass[letterpaper, 10 pt, conference]{ieeeconf}  

\IEEEoverridecommandlockouts                              

\usepackage{graphics} 
\usepackage{times} 
\usepackage{amsmath} 
\usepackage{amssymb}  
\usepackage{float}
\usepackage{mathtools}

\usepackage{caption}
\usepackage{graphicx}
\usepackage{multirow}
\usepackage{bm}
\usepackage{booktabs}
\usepackage{array}
\usepackage{gensymb} 
\usepackage{subcaption}
\usepackage{xcolor}
\usepackage[shortcuts,acronym]{glossaries}
\usepackage[symbol]{footmisc}
\usepackage{url}

\title{\LARGE \bf
\textsc{Kovis}: Keypoint-based Visual Servoing with Zero-Shot \\Sim-to-Real Transfer for Robotics Manipulation}

\author{En Yen Puang$^{1,2}$, Keng Peng Tee$^{1}$ and Wei Jing$^{1, 2, *}$ 
\thanks{$^{1}$ Institute for Infocomm Research (I$^2$R), A*STAR, Singapore}%
\thanks{$^{2}$ Institute for High Performance Computing (IHPC), A*STAR, Singapore}%
\thanks{{$^{*}$ Corresponding author, email addresses: \tt\small \{puangey, kptee, jingw\}@i2r.a-star.edu.sg}}%
}

\newacronym{vs}{VS}{Visual Servoing}
\newacronym{cnn}{CNN}{Convolutional Neural Networks}
\newacronym{fgsm}{FGSM}{Fast Gradient Sign Method}

\begin{document}

\maketitle
\thispagestyle{empty}
\pagestyle{empty}

\begin{abstract}
 
We present \textsc{Kovis}, a novel learning-based, calibration-free visual servoing method for fine robotic manipulation tasks with eye-in-hand stereo camera system. We train the deep neural network only in the simulated environment; and the trained model could be directly used for real-world visual servoing tasks. \textsc{Kovis} consists of two networks. The first keypoint network learns the keypoint representation from the image using with an autoencoder. Then the visual servoing network learns the motion based on keypoints extracted from the camera image. The two networks are trained end-to-end in the simulated environment by self-supervised learning without manual data labeling. After training with data augmentation, domain randomization, and adversarial examples, we are able to achieve zero-shot sim-to-real transfer to real-world robotic manipulation tasks. We demonstrate the effectiveness of the proposed method in both simulated environment and real-world experiment with different robotic manipulation tasks, including grasping, peg-in-hole insertion with 4mm clearance, and M13 screw insertion. The demo video is available at: \textit{http://youtu.be/gfBJBR2tDzA}
\end{abstract}

\section{Introduction}

\gls{vs} is a framework to control the motion based on the visual image input from the camera \cite{hutchinson1996tutorial, kragic2002survey}. \gls{vs} could be used to provide flexible vision-guided motion for many robotic manipulation tasks such as grasping, insertion, pushing. Therefore, for intelligent robotic applications in unstructured environment, \gls{vs} is highly desired for the robot to complete the manipulation tasks, by controlling the end-effector motion based on the visual feedback.

Traditionally, \gls{vs} usually requires extracting and tracking of visual features, in order to estimate the pose differences between the current pose and the target pose \cite{hutchinson1996tutorial, kragic2002survey, chaumette2016visual}. The pose difference will then be used as feedback for the \gls{vs} controller to move the robotic end-effector towards the target pose. The approach usually requires manually hand-crafted features for different applications, as well as manually labeling the target poses, which limits the task generalization of \gls{vs} in robotic applications. As a result of recent advancement in deep learning, making use of deep \gls{cnn} allows direct learning of the controller for \gls{vs}, instead of relying on manually handcraft features for the detection. Recent work on combining deep learning method with \gls{vs} has been explored by researchers \cite{yu2019siamese, bateux2018training}. However, these methods require collecting large amounts of training data from real experiments, and estimating camera pose differences based on input images. Thus, the target pose of the camera has to be identified for robotic manipulation. 

\begin{figure}[t]
    \centering
    \includegraphics[width=0.98\linewidth]{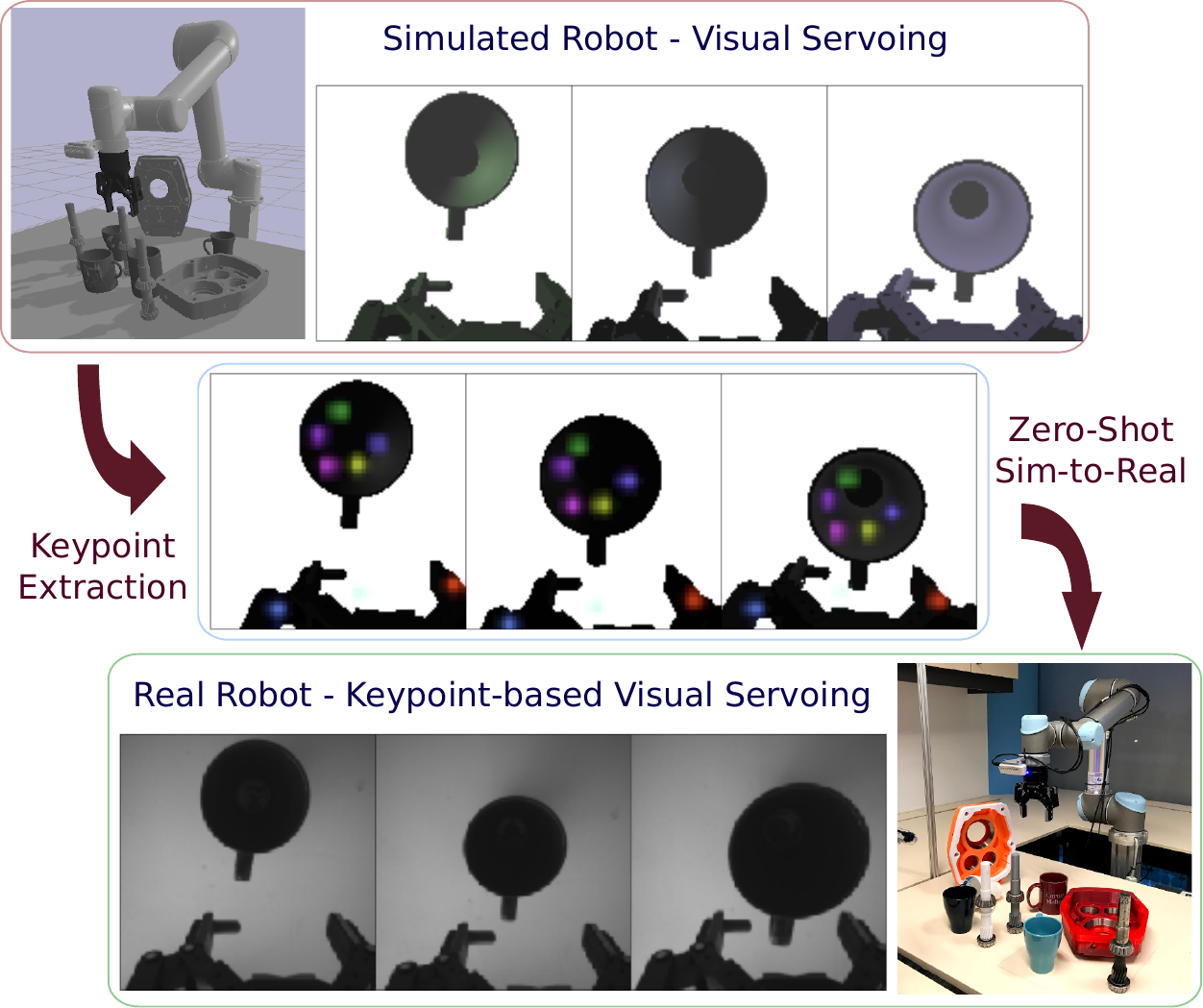}
    \caption{\textsc{Kovis} is a learning-based visual servoing framework for robotics manipulation tasks. It is trained entirely with synthetic data, and works on real-world scenarios with zero-shot transfer, using robust keypoints latent representation.}
    \label{fig:introduction}
\end{figure}

To achieve end-to-end, data-efficient deep learning based \gls{vs} method for fine robotic manipulation tasks, we propose \textsc{Kovis}, a \textbf{K}eyp\textbf{o}int based \textbf{Vi}sual \textbf{S}ervoing framework. \textsc{Kovis} learns the \gls{vs} controller that moves the robot end-effector to the target pose (e.g. pre-insertion pose) for manipulation tasks. The proposed \textsc{Kovis} learns the \gls{vs} only with synthesis data in simulated environment, and directly applies the learnt model in real-world robotic manipulator tasks. 

The main contributions of this paper are:
\begin{itemize}
    \item A general and efficient learning framework of \gls{vs} for robotic manipulation tasks;
    \item A self-supervised keypoint representation learning with autoencoder to identify the ``peg-and-hole'' relationship between the tool and target, thus achieves calibration-free in hand-in-eye system for the manipulation;
    \item An end-to-end, self-supervised learning method for \gls{vs} controller to move the robot end-effector to the target pose for manipulation without estimating the pose differences; and
    \item Combination of different training schemes to achieve zero-shot sim-to-real transfer for real-world fine manipulation tasks such as grasping and insertion.
\end{itemize}

\begin{figure*}
    \centering
    \includegraphics[width=0.98\linewidth]{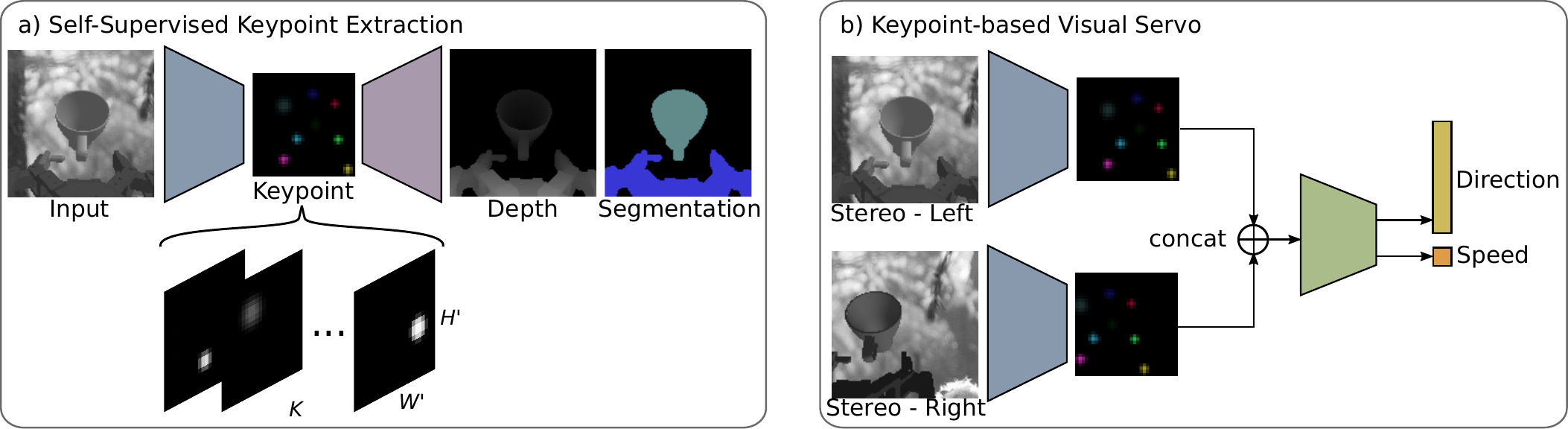}
    \captionsetup{format=hang}
    \caption{Architecture overview: \textsc{Kovis} consists of two major network modules that are trained together end-to-end: (a) Keypoint learning involves an autoencoder for self-supervision and a keypointer bottle-neck for robust latent representation; (b) Keypoint-based visual servo then predicts motion commands in terms of direction (unit-vector) and normalized speed (scalar) based-on the extracted keypoints from stereo inputs.}
    \label{fig:architecture}
\end{figure*}

\section{Relevant Work}

\subsection{Visual Perception and Representation for Manipulation}

Keypoints provide a compact representation for perception of images, and has been used in many computer vision applications such as image generation \cite{jakab2018unsupervised}, face recognition \cite{berretti20113d}, tracking \cite{hare2012efficient} and pose estimation \cite{tulsiani2015viewpoints}. Recently, keypoints have also been used as a representation for robotic manipulation tasks. For example, \cite{manuelli2019kpam, gao2019kpam} used keypoints for manipulation planning of category tasks with supervised learning, and \cite{qin2019keto} focused on learning a task-specific keypoint representation for robotic manipulation via self-supervised learning.

Autoencoders and their variations have also been applied in many robotic applications to deal with perception in manipulation \cite{byravan2018se3, puang2019visual}. An autoencoder maps the input image to a latent space, where the encoded latent value is usually used for planning \cite{qureshi2019motion} or as a feature extraction component for further learning process \cite{puang2019visual}.

Unlike supervised keypoint learning methods for robotic manipulation, which mostly focused on representation learning, our work utilizes \emph{self-supervised} learning with an autoencoder to extract the keypoints without any human labelling or annotations. In addition, our work includes learning \gls{vs} controller based on the learnt keypoint representation. 

\subsection{Visual Servoing}
Many research works have been conducted on \gls{vs} in past years. Traditionally, \gls{vs} algorithms usually rely on explicit geometric feature extraction \cite{cai2013uncalibrated, cai2016orthogonal}. Direct \gls{vs} methods have been explored, but they still require  global feature based image processing methods \cite{bateux2016histograms, bateux2016particle}. Recently, deep learning based methods have been used for \gls{vs}, including \gls{cnn} to learn the pose difference directly from the image input for the visual servoing \cite{bateux2018training, saxena2017exploring} and Siamese \gls{cnn} to learn the pose differences by comparing the difference of images of the target and current poses \cite{yu2019siamese}. 

The proposed \textsc{Kovis} directly learns the end-to-end robotic motion based on automatically-extracted keypoints of the image input, without estimating the pose difference. Our task and environment setting is similar to \cite{yu2019siamese}. However, our model is trained entirely in a simulated environment, without any data collected from real experiments.

\subsection{Self-supervised learning and sim-to-real transfer for robotic manipulation}

For robotic manipulation applications, it is expensive to get real-world data for learning-based methods. In order to overcome the data shortage problem in robotic learning applications, self-supervised learning in simulation environment \cite{fang2019learning}, and efficient sim-to-real transfer, are important \cite{tobin2017domain}. Several approaches have been proposed to address the sim-to-real problem, such as domain randomization  \cite{peng2018sim}, simulation randomization \cite{chebotar2019closing}, adversarial training \cite{zhang2019adversarial}. \textsc{Kovis} is data-efficient by adopting domain randomization and adversarial example techniques to achieve zero-shot sim-to-real transfer without requiring any real-world data.

\section{Method}
The proposed framework \textsc{Kovis}, as shown in Fig. \ref{fig:architecture}, consists of two major modules: an autoencoder for learning a keypoint representation from input images; and a feed-forward network for learning \gls{vs} motion commands. Both networks are trained end-to-end using ground-truth data gathered from simulations. To achieve zero-shot sim-to-real transfer, we also adapt several methods in the training to overcome the ``reality-gap'' and improve robustness in real-world manipulation scenarios.

\subsection{Self-Supervised Keypoint Extraction}
\label{sec:keypoint}
We first learn the latent keypoint representation of objects in a \gls{vs} scene with an \gls{cnn}-based autoencoder. For an input image $\boldsymbol{x} \in \mathbb{R}^{H \times W \times C}$, we formulate the keypoint $\boldsymbol{k} \in \mathbb{R}^{H' \times W' \times K}$ as a latent representation in the autoencoder architecture where the encoder $\mathnormal{f}: \boldsymbol{x} \mapsto \boldsymbol{z}$, keypointer $\Phi: \boldsymbol{z} \mapsto \boldsymbol{k}$ and decoder $\mathnormal{g}: \boldsymbol{k} \mapsto \boldsymbol{x}'$ are optimized to minimize reconstruction loss in self-supervised approach:
\begin{align} \label{eqn:min_recon_loss}
\min_{\theta_f, \theta_g} \|\boldsymbol{x} - \left(g\ast\Phi\ast f\right)\left(\boldsymbol{x}\right)\|_2
\end{align} 

Succeeding the encoder, the keypointer $\Phi$ transforms the feature map $\boldsymbol{z}$ into $K$ individual keypoints $\boldsymbol{k}_i$ on the 2D feature maps $\Omega = \mathbb{Z}^{H' \times W'}$ in two steps. First, the softmax of $\boldsymbol{z}$ is used for computing the channel-wise 2D centroid for each channel. Then a 2D Gaussian distribution with fixed covariance $\rho^{-1}$ is used to model the unimodal Gaussian keypoint with mean the centroid of the channel's softmax:
\begin{align} \label{eqn:activation}
\boldsymbol{j}^*_i &= \sum_{\boldsymbol{j} \in \Omega} \boldsymbol{j} \frac{ \exp\left(\boldsymbol{z}_i\right)}{\sum_\Omega \exp\left(\boldsymbol{z}_i\right)} \\
\alpha_i &= \sigma\left(\max_\Omega\left(\boldsymbol{z}_i\right)\right)
\end{align}
where $\boldsymbol{j} = (j_1$, $j_2)$ represent the indices in vertical and horizontal axes in $\Omega$, and $\boldsymbol{j}^*$ the centroid of the 2D feature map. This keypoint formulation is similar to \cite{jakab2018unsupervised} but with additional keypoint confidence $\alpha$ from the sigmoid $\sigma(\cdot)$ of the channel's maximum activation:
\begin{align} \label{eqn:keypoints}
\boldsymbol{k}_i = \alpha_i \prod_{j=(j_1, j_2)} \exp\left(\text{-}\rho \left(j - j^*_i\right)^2\right) \quad \forall \boldsymbol{j} \in \Omega
\end{align}

\begin{figure}
    \centering
    \includegraphics[width=1\linewidth]{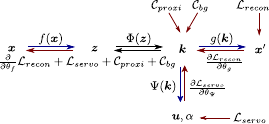}
    \caption{Training of \textsc{Kovis}: All components are trained end-to-end together with the differentiable keypointer $\Phi$ which contain no learnable parameters.}
    \label{fig:gradient}
\end{figure}

Additionally, we enforce two soft constraints in the training of keypoints extraction to achieve better feature localization and representation. Since widely-spread keypoints are more distinct compared to concentrated ones, the first constraint $\mathcal{C}_{proxi} $ encourages better representation by pushing keypoints away from each other through penalizing the L2-norm among extracted keypoints centroid as in \cite{zhang2018unsupervised} with a hyper-parameter $\gamma$:
\begin{align} \label{eqn:loss_proxi}
\mathcal{C}_{proxi} = \sum^{K}_{i,i'\mid i'>i} \alpha_i \alpha_{i'} \exp\left(\text{-}\gamma\|\boldsymbol{j}^*_i - \boldsymbol{j}^*_{i'}\|_2\right)
\end{align}

To prevent arbitrary keypoint formation which is bad for interpretability, the second constraint $\mathcal{C}_{bg}$ encourages better features localization by penalising any keypoints that fall into the background segmentation mask of the input image:
\begin{align} \label{eqn:loss_bg}
\mathcal{C}_{bg} = \sum_i^K \sum_{\boldsymbol{j}}^\Omega \mathbb{I}\left[\boldsymbol{j}\in B\right]\boldsymbol{k}_{i\boldsymbol{j}}
\end{align}
where $\mathbb{I}\left[\boldsymbol{j}\in B\right]$ is a binary logic that returns true when $\boldsymbol{j}$ is in the background set $B$ of ground-truth segmentation mask (assumed to be unavailable during inference). Both constraints are scaled by a keypoint confidence $\alpha$ to reduce contributions from low-confidence keypoints. The combination of these 2 constraints ensure that the extracted keypoints are localized within object of interest, and hence perform better at capturing essential information regarding object geometry for the downstream \gls{vs} task.

\subsection{Self-Supervised Keypoint-based Visual Servoing}
\label{sec:servo}

\gls{vs} is a type of reactive controller whose objective is to generate motion that minimizes the differences between the goal and current visual observation/feedback. In contrast to conventional \gls{vs} in which the servo takes in both current and goal states, \textsc{Kovis} trains a servo for a single task and hence only requires the current state as input during inference. Given the extracted keypoints $\boldsymbol{k}$ from input image, the servo $\Psi: \boldsymbol{k} \mapsto (\boldsymbol{u}, \beta)$ is a multi-layer Fully-Connected network trained with supervised learning approach to predict the motion of the robot end-effector. The predicted motion consists of the direction $\boldsymbol{u} \in \mathbb{R}^d \mid |\boldsymbol{u}| \coloneqq 1$ of the end-effector, as well as its normalized speed $\beta \in [0, 1]$ which is 0 at the desired pose and saturates to 1 when far away. 

\begin{figure}
    \centering
    \includegraphics[width=0.8\linewidth]{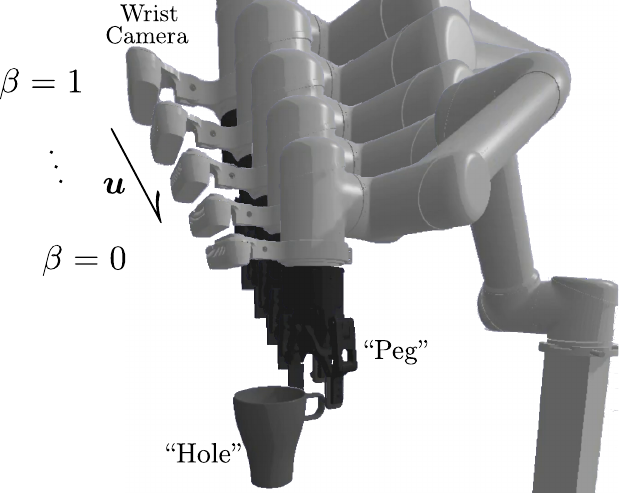}
    \caption{Simulating visual servoing task with UR5 robot, 2-finger gripper and a wrist camera. For mug picking task, gripper is the ``peg'' and mug handle is the target ``hole''.}
    \label{fig:simulation}
\end{figure}

This setup is not to align keypoints between current and goal states but to directly minimize the differences between the predicted and ground-truth motion. In other words, the servoing task is determined by the training data as depicted in Fig. \ref{fig:simulation}. We define the loss function of the servo as:
\begin{align} \label{eqn:loss_servo}
\mathcal{L}_{servo} = \beta^*\left(1-\frac{\boldsymbol{u}^\intercal\boldsymbol{u}^*}{|\boldsymbol{u}|}\right)+\text{BCE}(\beta, \beta^*)
\end{align}
which consists of a scaled inverted cosine similarity for $\boldsymbol{u}$ and binary-cross-entropy loss BCE$(\cdot)$ for $\beta$ with $\boldsymbol{u}^*$ and $\beta^*$ as the ground-truth. Note that the former is scaled by $\beta^*$ to reduces the loss from direction when the speed is low. 

\subsection{End-to-End Training and Zero-Shot Transfer}
\label{sec:end2end}

\textsc{Kovis} is trained end-to-end using the input images from eye-in-hand stereo camera to the motion commands in robot end-effector frame. The training is done entirely on synthetic data gathered from a simulated environment. 

{\noindent\bf Training Data Generation.}
The first step for this framework therefore is to setup the simulated \gls{vs} task for generating training data. This involves 1.) getting the CAD models of the involved objects, 2.) defining their target pose and motion for each of the components, and lastly 3.) the robotic and camera system according to the requirement of the intended manipulation task. Once the \gls{vs} task is established, synthetic images paired with motion ground-truth are generated and then used for the training in Sections \ref{sec:keypoint} and \ref{sec:servo}. Data are recorded in the form of roll-outs. Each roll-out starts at the same target pose and then propagates out in the opposite of $\boldsymbol{u}$ with a speed factor, both randomized among roll-outs. Roll-out terminates early when collision occurs.

{\noindent\bf Object Anchoring.}
Motion commands are generated in end-effector frame based on images from a camera whose transformation relative to the end-effector is not calibrated. In other words, \textsc{Kovis} realizes camera calibration-free configuration for manipulative \gls{vs} by identifying ``peg-AND-hole'' relationship from the input image during training. This relationship requires the ``hole'' (i.e.\ normally the target object) to act as the pose anchor while the servo generates the motion command for the ``peg'' (e.g.\ gripper or tool) based on the relative pose between them, instead of the absolute camera-object pose that rely on deliberate camera calibrations. Moreover, the dimension of motion direction $\boldsymbol{u}$ is adapted accordingly to disambiguate symmetries in object geometry, as well as the requirement of the \gls{vs} task itself.

{\noindent\bf Data Domain and Losses.}
``Reality-gap'' \cite{tobin2017domain} is a problem for models trained entirely in simulation without any real world adaptation. In staying agnostic towards object texture, the input image $\boldsymbol{x}$ is transformed into grayscale domain, and the decoder's outputs $\boldsymbol{x}'\coloneqq\left(\boldsymbol{x}^d, \boldsymbol{x}^s\right)$ reconstructs the depth buffer $\boldsymbol{x}^d$ and semantic segmentation $\boldsymbol{x}^s$ only using the keypoint. This is to ensure that the keypoint encodes only the essential geometric information and not overfit subtle details or artifacts in the simulation.
In staying agnostic towards the quality of real depth images, keypoints extracted from stereo pair images are used instead in training the servo by concatenating 2 sets of keypoints extracted individually from the left and right stereo images. As the result, without loss of generality, this setup assumed the use of stereo camera in the intended \gls{vs} task.

The total loss function $\mathcal{L}$ for training \textsc{Kovis} is the sum of all losses and constraints:
\begin{align} \label{eqn:losses}
    \mathcal{L} &= \mathcal{L}_{recon} + \mathcal{L}_{servo} + \mathcal{C}_{proxi} + \mathcal{C}_{bg}
\end{align}
where $\mathcal{L}_{recon}$ is the reconstruction loss which involve mean-squared-error for depth buffer and multi-class cross-entropy for semantic segmentation for both left and right stereo input images. Fig. \ref{fig:gradient} depicts the semantics of end-to-end training of all components in \textsc{Kovis}.

{\noindent\bf Adversarial Examples and Training.}
We adopt image augmentation (e.g.\ lighting, blurring and shifting) and domain randomization (background, texture, camera pose) \cite{tobin2017domain} methods for the training in simulated environment. In addition, adversarial examples are used to effectively augment training data which the hand-crafted methods could not achieve \cite{xie2019adversarial}. A randomized combination of multiple generation methods including \gls{fgsm}, iterative \gls{fgsm} and least-likely \gls{fgsm} are used together with random augmentation strength and number of iterations as in \cite{puang2019visual}. These augmentations are applied to the entire mini-batch before being used for training. This method, hereinafter referred to as Adex, aims to widen the domain that the encoder is able to operate in, thereby becoming more robust when handling real-world data, despite being trained only on synthetic ones. 

\section{Implementations and Experiments}

\begin{figure}[t]
    \centering
    \includegraphics[width=1\linewidth]{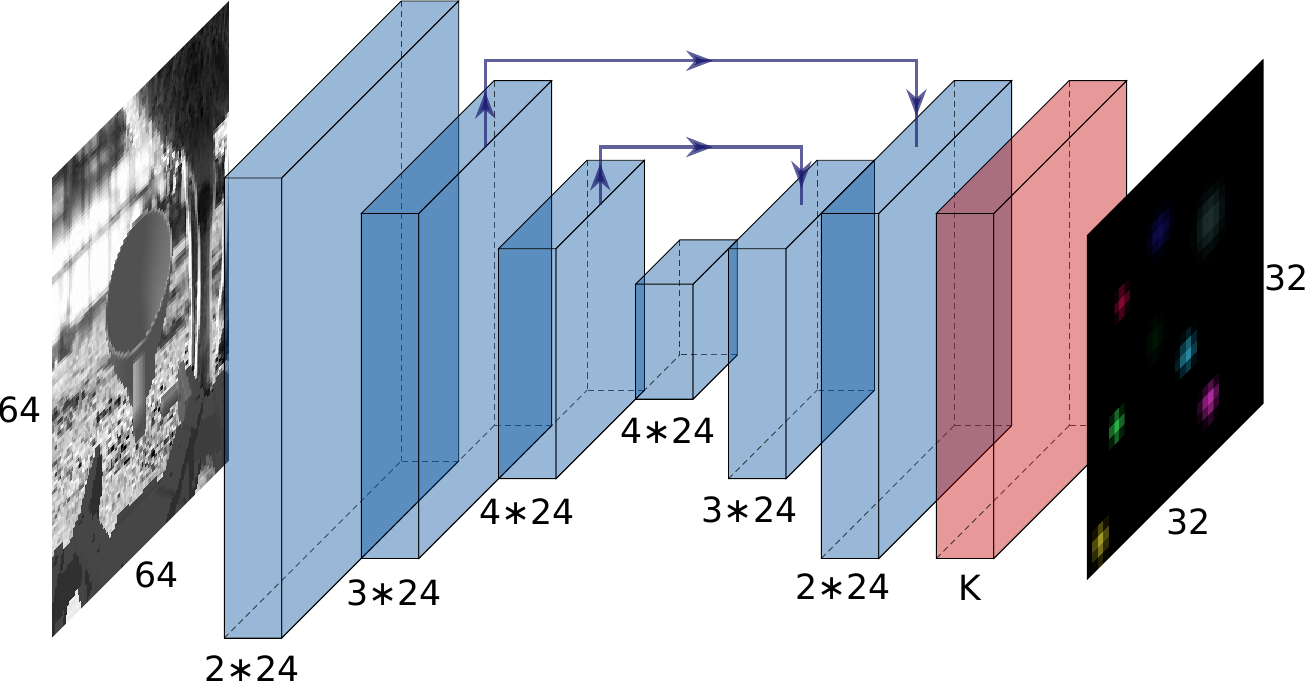}
    \caption{The encoder (blue) is a U-Net with DenseNet backbone which has total 0.5M parameters. It consists of several DenseBlocks each with 2 to 4 DenseLayers and a growth-rate of 24. Output of keypointer (red) is a 32 $\times$ 32 ($H' \times W'$) feature map with $K$ channels representing $K$ keypoints.}
    \label{fig:encoder}
\end{figure}

\begin{figure}[t]
    \centering
    \includegraphics[width=1.0\linewidth]{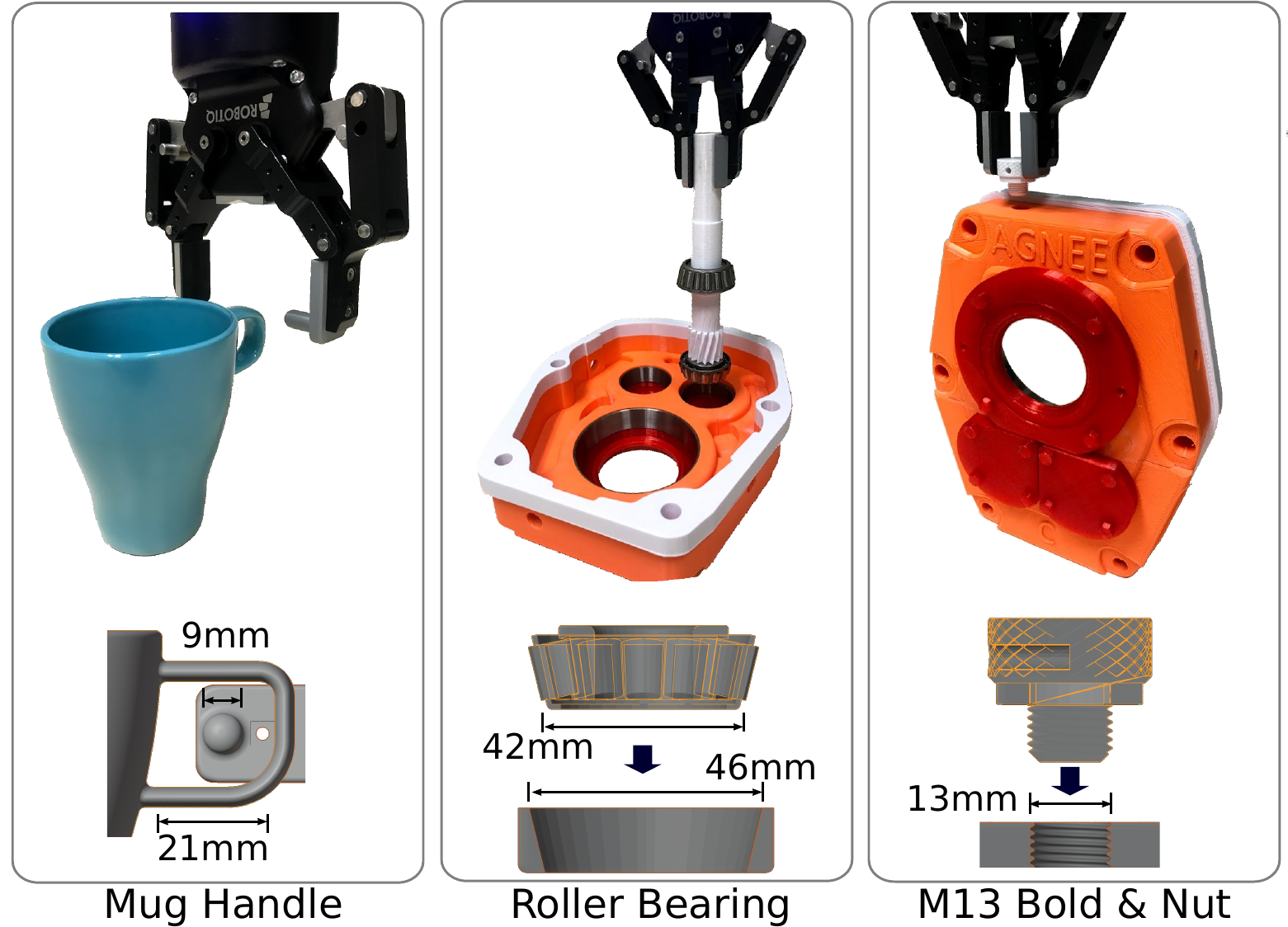}
    \caption{Visual servoing for pick-and-place alignment tasks: Pick-Mug (left), Insert-Shaft (mid) and Insert-Plug (right).}
    \label{fig:alignment}
\end{figure}

\begin{figure*}[t]
    \centering
    \includegraphics[width=1.0\linewidth]{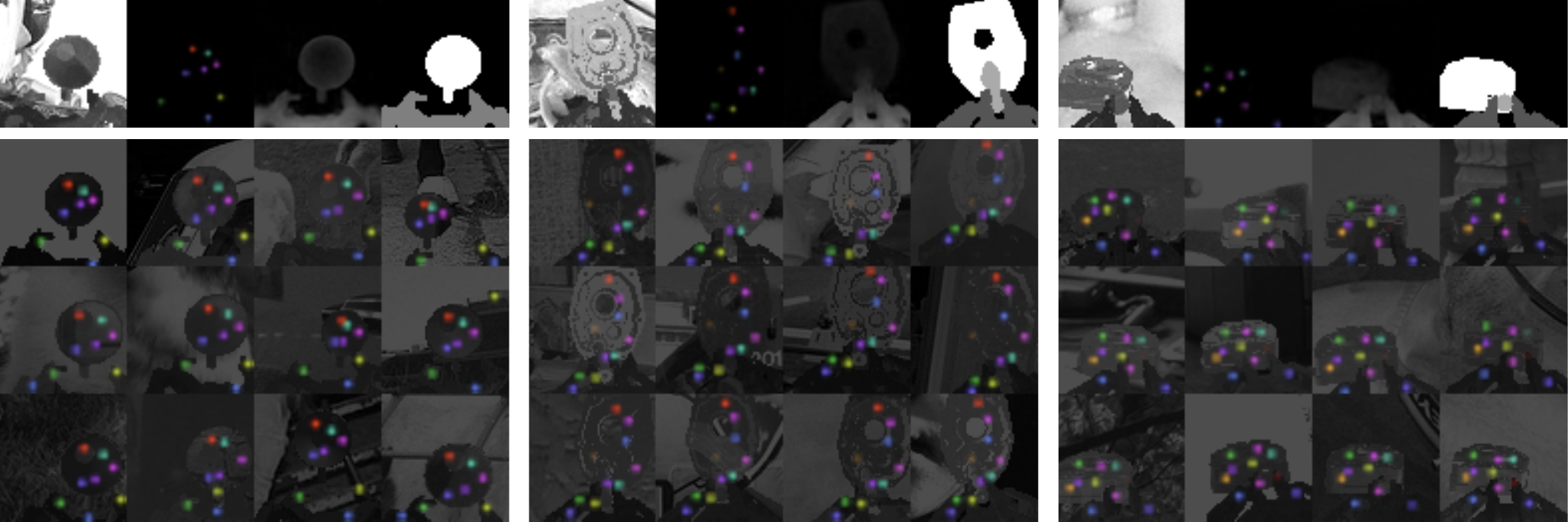}
    \caption{Top row: Synthetic input image, extracted keypoint, predicted depth buffer and semantic segmentation for 3 \gls{vs} tasks. Bottom row: Overlaid images between inputs and their respective extracted keypoints for the respectively \gls{vs} tasks.}
    \label{fig:outputs}
\end{figure*}

\begin{table*}
    \centering
    \def\arraystretch{1.2}
    \caption{Ablation study on the effect of input image perturbations on keypoint localization. Average difference of keypoint location and confidence $(\delta\boldsymbol{j^*}, \delta\alpha)$ (the lower the better) between none and various perturbations are shown here. Keypoint location are normalized from $[0, H']$ and $[0, W']$ to $[0, 1]$ while keypoint confidence remains between $[0, 1]$ in percentage $\%$.}
    \begin{tabular}{ccccccccc}
    \toprule
         \multirow{2}{*}{Background} & \multirow{2}{*}{Texture} & \multirow{2}{*}{Lighting} & \multicolumn{2}{c}{Pick-Mug} & \multicolumn{2}{c}{Insert-Shaft} & \multicolumn{2}{c}{Insert-Plug}  \\
         &&& w/o. Adex & with Adex & w/o. Adex & with Adex & w/o. Adex & with Adex \\
    \midrule
         \checkmark&&&                    (0.5, 0.7) & (0.5, 0.6) &  (0.2, 0.1) & (0.2, 0.0) &  (0.8, 1.2) & (1.1, 0.5) \\
         &\checkmark&&                    (0.4, 0.5) & (0.4, 0.5) &  (0.2, 0.0) & (0.2, 0.0) &  (0.5, 1.0) & (0.5, 0.3) \\
         &&\checkmark&                    (0.3, 0.4) & (0.3, 0.4) &  (0.3, 0.1) & (0.4, 0.0) &  (0.7, 1.1) & (0.8, 0.4) \\
         \checkmark&\checkmark&\checkmark&(0.6, 0.7) & (0.5, 0.6) &  (0.3, 0.1) & (0.4, 0.0) &  (0.9, 1.2) & (1.2, 0.5) \\
    \bottomrule
        & Average && 0.51 & \textbf{0.48} & 0.16 & \textbf{0.15} & 0.93 & \textbf{0.66}
    \end{tabular}
    \label{tab:perturbIm}
\end{table*}

We implement \textsc{Kovis} as the ``last-mile'' solution for robotics manipulation tasks. Here we demonstrate 3 applications on \gls{vs} alignment for pick-and-place tasks with an eye-in-hand stereo camera mounted near robot end-effector as depicted in Fig. \ref{fig:simulation} and Fig. \ref{fig:alignment}.

The encoder is a Fully-Convolutional U-Net \cite{ronneberger2015u} based on DenseNet \cite{huang2017densely} architecture with skip-connections (by concatenation) linking downward and upward streams, as shown in Fig. \ref{fig:encoder}. The decoder on the other hand is a feed-forward expanding DenseNets while the servo is made of multiple Fully-Connected layers. All DenseNet used here are of \textit{Weight-BatchNorm-Activation} form and all convolutional layers has increasing dilation as in \cite{puang2019visual}. The total learnable parameters of \textsc{Kovis} is around 1M. 

The settings of hyper-parameters highly depend on the application scenarios. For applications that require high precision, higher $\rho$ (Eq. \ref{eqn:keypoints}), $H$ and $W$ (Eq. \ref{eqn:min_recon_loss}) are preferred. For applications that involve complex or large object geometry more keypoint $K$ (Eq. \ref{eqn:min_recon_loss}) and smaller $\gamma$ (Eq.  \ref{eqn:loss_proxi}) are preferred. In all of our experiments we use a fixed combination of $64\times 64$ ($H\times W$) input image size, $32\times 32$ ($H'\times W'$) keypoint spatial size, 2.5 for $\rho$ in keypoint representation and 20 for $\gamma$ in proximity constraint.

For the experiment setup, we use a UR5 robotic arm, a Robotiq 2F-85 gripper, and a Intel Realsense D435 camera (which we only use its stereo images) mounted on the robot arm. In the simulation environment \cite{coumans2019}, we create a virtual robot workspace by duplicating our real-world robotics system. For each of the \gls{vs} task, we simulate 7000 servoing roll-outs and collected around 35000 set of images (i.e.\ color, depth and segmentation mask) paired with the ground-truth motion commands in the end-effector frame. 

\subsection{Keypoint Localization}
In this experiment we evaluate the localization capability of the learnt keypoint and its robustness under input perturbations. First, we show the extracted keypoints from various viewing angles of the three \gls{vs} tasks. Fig. \ref{fig:outputs} shows that keypoints consistently locate object of interest (``peg'' and ``'hole') in each of the tasks. Note that in the absence of severely challenging background, none of the keypoints are located outside of object of interest, and their formation change accordingly to its input. These show that the keypoint extraction encodes the correct object geometric information  instead of just any arbitrary latent representation. Moreover, from Insert-Plug task ($K=10$) we observe that excessive keypoints are automatically trimmed (low $\alpha_i$) thanks to the soft constraints and keypoint confidence used in \textsc{Kovis}.

\begin{table*}
    \centering
    \def\arraystretch{1.2}
    \caption{Servo consistency in term of scaled inverted cosine similarity loss for direction and BCE loss for speed $(\mathcal{L}_{\boldsymbol{u}}, \mathcal{L}_\beta)$ (the lower the better) over camera perturbations at different magnitudes (3D translations, 3D rotations).}
    \begin{tabular}{ccccccc}
    \toprule
         \multirow{2}{*}{Camera Perturbation} & \multicolumn{2}{c}{Pick-Mug} & \multicolumn{2}{c}{Insert-Shaft} & \multicolumn{2}{c}{Insert-Plug}  \\
         & w/o. Adex & with Adex & w/o. Adex & with Adex & w/o. Adex & with Adex \\
    \midrule
         ($\pm$0.5cm, $\pm2.5^\circ$) & (0.022, 0.153) & (0.021, 0.161) &  (0.018, 0.165) & (0.014, 0.161) &  (0.039, 0.173) & (0.031, 0.174) \\
         ($\pm$1.0cm, $\pm5.0^\circ$) & (0.024, 0.167) & (0.028, 0.167) &  (0.024, 0.178) & (0.018, 0.172) &  (0.043, 0.183) & (0.035, 0.183) \\
         ($\pm$1.5cm, $\pm7.5^\circ$) & (0.039, 0.177) & (0.042, 0.182) &  (0.041, 0.215) & (0.039, 0.213) &  (0.047, 0.214) & (0.043, 0.213) \\
    \bottomrule
        Average & (\textbf{0.028, 0.166}) & (0.033, 0.170) & (0.028, 0.186) & (\textbf{0.024}, \textbf{0.182}) & (0.043, 0.190) & (\textbf{0.036}, \textbf{0.190})
    \end{tabular}
    \label{tab:perturbCam}
\end{table*}

Next, we evaluate keypoint stability under input image perturbations. Table \ref{tab:perturbIm} shows the ablation study on the effect of keypoint localization stability over input image perturbations. The differences of keypoint location $\delta\boldsymbol{j^*}$ and confidence $\delta\alpha$ are compared when background, texture and lighting perturbations are introduced , while camera pose perturbation is disabled. More than 400 synthetic input images from each tasks are sampled and the resulted keypoints after the perturbations are compared to itself before any perturbations. As shown in Table \ref{tab:perturbIm}, Adex generally reduces the changes in keypoint confidence but introduces noise in keypoint location. Moreover, random background changes has the highest influences in keypoint localization than the others. Stability of keypoint localization of objects, nonetheless, is robust against input perturbations as the fluctuations only appeared to be at the scale of $0.1\%$. 

Lastly we evaluate the performance of object anchoring through camera pose perturbations. During synthetic training data generation the 6D camera pose is perturbed ($\pm$1cm, $\pm$5$^\circ$) for every image captured to encourage object anchoring during training, described in Section \ref{sec:end2end}. Here we perturb the camera pose with different magnitude and observe the error (with respect to its ground-truth) of the predicted motion commands in term of scaled inverted cosine similarity loss for direction $\mathcal{L}_{\boldsymbol{u}}$ and BCE loss for speed $\mathcal{L}_\beta$ shown in Eq. \ref{eqn:loss_servo}. As shown in Table \ref{tab:perturbCam} quality of motion commands deteriorates as the perturbation increases. On the other hand, performance is improved by Adex and \textsc{Kovis} is able to maintain robustness even when magnitude of camera pose perturbations is higher than how it was trained.

\subsection{Alignment with Eye-in-hand Camera}
 In this experiment we evaluate \textsc{Kovis} in real world \gls{vs} manipulation tasks. Tasks depicted in Fig. \ref{fig:alignment} are different in their object size and margin of error. The first task is a mug picking task in which the robot is required to place its hook into the mug's handle as the gripper closes. The second and third task are insert-task which involve a shaft on tapered roller bearing and a plug on M13 screw respectively. The robot is required to orient the ``peg'' with the ``hole'' before a downward push to complete the insertion. All tasks are being trained with 4 dimensional motion ($d=4$ in Eq. \ref{eqn:loss_servo}) namely $xyz$ translation and  $xy$ in-plane rotation, and with 5, 10 and 16 number of keypoint. Table \ref{tab:alignment} shows the success rate under complex background over 10 runs each. Success rate is hindered by sub-optimal number of keypoint that causes under or over-fitting. Besides, Adex improves the performance only when the number of keypoint is appropriate. In general, \textsc{Kovis} is able to complete these 3 servo tasks with more than 90\% success rate using Adex and 10 keypoints.

\begin{figure*}
    \centering
    \includegraphics[width=1.0\linewidth]{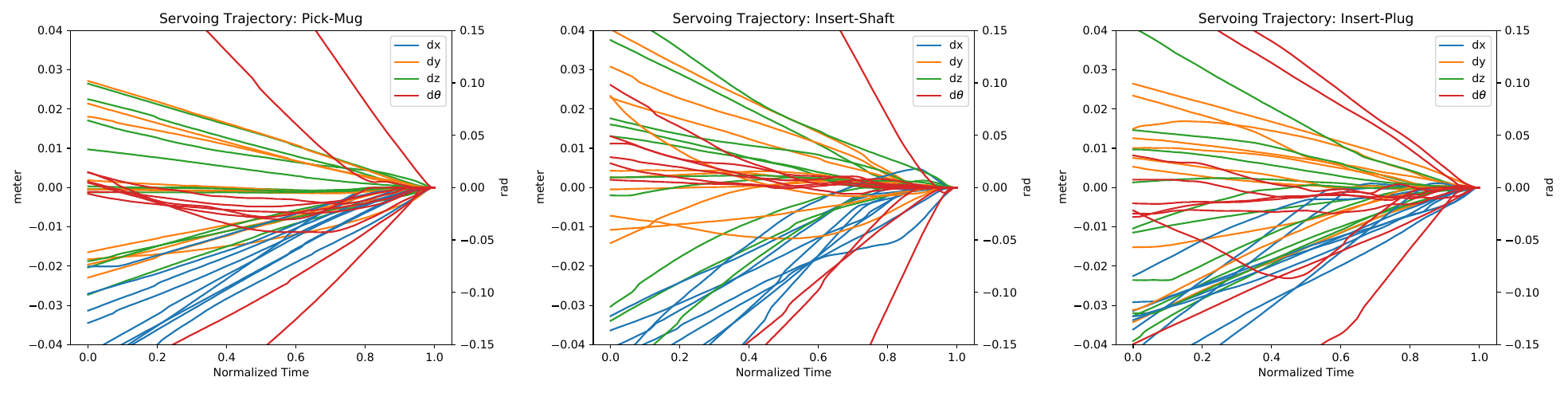}
    \caption{Trajectories of robot end-effector in 4 dimensional pose against normalized time for each of the three \gls{vs} tasks.}
    \label{fig:trajectory}
\end{figure*}

Besides accuracy, we show the servoing smoothness by tracking the robot end-effector trajectory from multiple initial poses. Fig \ref{fig:trajectory} depicts 10 trajectories for each of the tasks. Error of end-effector pose in 4-dimensional from the goal pose are plotted against normalized time (stretched to accommodate trajectories with various length). Trajectories are generally smooth while following the predicted direction $\boldsymbol{u}$ with small magnitude linearly proportional to $\beta$, partly due to the nature of speed control at robot end-effector. As the result each trajectory takes about 6 seconds to complete depending on the initial pose. 

\begin{table}
    \centering
    \def\arraystretch{1.2}
    \caption{Success rate of \textsc{Kovis} on three real world \gls{vs} tasks under complex background with multiple number of keypoint and use of Adex.}
    \begin{tabular}{rccc}
    \toprule
        Servo Task & $K=5$ & $K$ = 10 & $K$ = 16 \\
    \midrule
        \multicolumn{1}{l}{Pick-Mug} &&&\\
        w/o. Adex & 1.0 & 1.0 & 1.0 \\
        with Adex & 1.0 & 1.0 & 1.0 \\
        \hline
        \multicolumn{1}{l}{Insert-Shaft} &&&\\
        w/o. Adex & 0.2 & 0.6 & 0.6  \\
        with Adex & 0.0 & \textbf{1.0} & 0.6  \\
        \hline
        \multicolumn{1}{l}{Insert-Plug} &&&\\
        w/o. Adex & 0.8 & 0.8 & 0.9   \\
        with Adex & 0.6 & \textbf{0.9} & 0.2  \\
    \bottomrule
    \end{tabular}
    \label{tab:alignment}
\end{table}

\section{Conclusion}

In this paper, we presented \textsc{Kovis}, a \textbf{K}eyp\textbf{o}int based \textbf{Vi}sual \textbf{S}ervoing framework for the "last-mile" robotic manipulation task. 
\textsc{Kovis} learns an efficient and effective keypoint representation for identifying object geometric information in robotic \gls{vs} settings. 
It consists two major modules, an autoencoder for keypoint extraction, and a \gls{vs} network for learning the robotic motion. Both networks are trained end-to-end and entirely on synthetic data. \textsc{Kovis} does not require any real-world data or adaptation and achieves zero-shot sim-to-real transfer by having multiple data augmentations strategies. 
In addition, external calibration of hand-in-eye camera is not required for manipulation tasks due to \textsc{Kovis}'s end-to-end nature (input image to motion command) and its ability to identify the geometric ``peg-and-hole'' relationships between the tool and the target.
The effectiveness of the proposed \textsc{Kovis} has been demonstrated in several fine robotic manipulation tasks with high success rate. Through experiments we also demonstrate the stability the predicted keypoints and motion commands over input image and camera pose perturbations. 
The future work will be on extending the current methods to category-level generalization for robotic manipulation tasks, as well as using multi-modality sensory information to achieve better performance on fine manipulation tasks.

\section*{Acknowledgement}

\noindent This research is supported by the Agency for Science, Technology and Research (A*STAR), Singapore, under its AME Programmatic Funding Scheme (Project \#A18A2b0046).


\bibliography{2020_visual_servo}

\end{document}